%% file: conference_101719.tex
\documentclass[conference]{IEEEtran}
\IEEEoverridecommandlockouts
\usepackage{cite}
\usepackage{amsmath,amssymb,amsfonts}
\usepackage{algorithmic}
\usepackage{graphicx}
\usepackage{textcomp}
\usepackage{xcolor}
\usepackage{tikz}
\usepackage[xindy,acronym]{glossaries}
\usetikzlibrary{trees}
\usepackage{verbatim}
\usepackage{tablefootnote}

\def\BibTeX{{\rm B\kern-.05em{\sc i\kern-.025em b}\kern-.08em
    T\kern-.1667em\lower.7ex\hbox{E}\kern-.125emX}}
    
\begin{document}

\input{abbreviations}

\title{A Survey on the Integration of Generative AI for Critical Thinking in Mobile Networks}

\author{\IEEEauthorblockN{Athanasios Karapantelakis, Alexandros Nikou, Ajay Kattepur, Jean Martins, Leonid Mokrushin, \\Swarup Kumar Mohalik, Marin Orlic, Aneta Vulgarakis Feljan}
\IEEEauthorblockA{\textit{Ericsson Research} \\
firstname.lastname@ericsson.com}}

\maketitle

\begin{abstract}
In the near future, mobile networks are expected to broaden their services and coverage to accommodate a larger user base and diverse user needs. Thus, they will increasingly rely on artificial intelligence (AI) to manage network operation and control costs, undertaking complex decision-making roles. This shift will necessitate the application of techniques that incorporate critical thinking abilities, including reasoning and planning. Symbolic AI techniques already facilitate critical thinking based on existing knowledge. Yet, their use in telecommunications is hindered by the high cost of mostly manual curation of this knowledge and high computational complexity of reasoning tasks. 
At the same time, there is a spurt of innovations in industries such as telecommunications due to Generative AI (GenAI) technologies, operating independently of human-curated knowledge. However, their capacity for critical thinking remains uncertain. This paper aims to address this gap by examining the current status of GenAI algorithms with critical thinking capabilities and investigating their potential applications in telecom networks. Specifically, the aim of this study is to offer an introduction to the potential utilization of GenAI for critical thinking techniques in mobile networks, while also establishing a foundation for future research.
\end{abstract}


\input{sections/section1_introduction}
\input{sections/section2_background}

\input{sections/section3_relevant_approaches}

\input{sections/section4_telco-uc}
\input{sections/section5_challenges_opportunities}
\input{sections/section6_conclusion}

\bibliographystyle{ieeetr}
\bibliography{references}
\end{document}

%% file: abbreviations.tex

\newacronym{TV}{TV}{television}
\newacronym{DTTB}{DTTB}{digital television terrestrial broadcasting}
\newacronym{DVB}{DVB}{Digital Video Broadcast}
\newacronym{DVB-H}{DVB-H}{Digital Video Broadcast-Handheld}
\newacronym{ATSC}{ATSC}{Advanced Television System Committee}
\newacronym{ATSC-M/H}{ATSC-M/H}{Advanced Television System Committee - Mobile/Handheld}
\newacronym{IPTV}{IPTV}{Internet Protocol television}
\newacronym{IP}{IP}{Internet Protocol}
\newacronym{l1}{L1}{Layer 1}
\newacronym{l2}{L2}{Layer 2}
\newacronym{l3}{L3}{Layer 3}
\newacronym{cnn}{CNN}{Convolutional Neural Network}
\newacronym{ann}{ANN}{Artificial Neural Network}
\newacronym{scef}{SCEF}{Service Capability Exposure Function}
\newacronym{zipnet}{ZipNet}{Zipper Network}
\newacronym{rrm}{RRM}{Radio Resource Management}
\newacronym{fps}{FPS}{Frames Per Second}
\newacronym{qa}{QA}{Question Answering}
\newacronym{ar}{AR}{Augmeneted Reality}

\newacronym{cdl}{CDL}{Clustered Delay Line}
\newacronym{tdl}{TDL}{Tapped Delay Line}

\newacronym{HR}{HR}{Human Resources}

\newacronym{tot}{ToT}{Tree-of-Thoughts}
\newacronym{got}{GoT}{Graph-of-Thoughts}

\newacronym{DGM}{DGM}{Deep Generative Model}
\newacronym{PGM}{PGM}{Probabilistic Graphical Model}
\newacronym{VAE}{VAE}{Variational autoencoder}
\newacronym{ARM}{ARM}{Autoregressive model}
\newacronym{NF}{NF}{Normalizing flows}
\newacronym{PDDL}{PDDL}{Planning Domain Definition Language}
\newacronym{RL}{RL}{Reinforcement Learning}
\newacronym{react}{ReAct}{Reasoning and Acting}
\newacronym{pal}{PAL}{Program-aided Language Models}

\newacronym{CSI}{CSI}{Channel State Information}
\newacronym{ml}{ML}{Machine Learning}
\newacronym{bss}{BSS}{Business Support System}
\newacronym{nlp}{NLP}{Natural Language Processing}
\newacronym{1g}{1G}{first generation of mobile networks}
\newacronym{2g}{2G}{second generation of mobile networks}
\newacronym{2.5g}{2.5G}{Transitional 2.5 generation of mobile networks}
\newacronym{3g}{3G}{third generation of mobile networks}
\newacronym{4g}{4G}{fourth generation of mobile networks}
\newacronym{5G}{5G}{fifth generation of mobile networks}
\newacronym{5g}{5G}{fifth generation of mobile networks}
\newacronym{6g}{6G}{sixth generation of mobile networks}
\newacronym{nlg}{NLG}{natural language generation}
\newacronym
[
  longplural={Large Language Models}
]
{llm}{LLM}{Large Language Model}
\newacronym
[
  longplural={Bidirectional Encoder Representations from Transformers}
]
{bert}{BERT}{Bidirectional Encoder Representations from Transformer}
\newacronym
[
  longplural={Universal Sentence Encoder}
]
{use}{USE}{Universal Sentence Encoder}
\newacronym
[
  longplural={Radio Base Stations}
]
{rbs}{RBS}{Radio Base Station}
\newacronym
[
  longplural={Natural Language Understanding}
]
{nlu}{NLU}{Natural Language Understanding}
\newacronym
[
  longplural={Network Functions}
]
{nf}{NF}{Network Function}
\newacronym
[
  longplural={Generative Pretrained Transformers}
]
{gpt}{GPT}{Generative Pretrained Transformer}

\newacronym
[
  longplural={Knowledge Bases}
]
{kb}{KB}{Knowledge Base}

\newacronym{palm}{PaLM}{Pathways Language Model}
\newacronym{llama}{LLAMA}{Large Language Model Meta AI}

\newacronym{cot}{CoT}{Chain-of-Thought}

\newacronym
[
  longplural={Generative Pretrained Transformers}
]
{rouge}{ROUGE}{Recall-Oriented Understudy for Gisting Evaluation}
\newacronym
[
  longplural={Deep Belief Networks}
]
{dbn}{DBN}{Deep Belief Network}
\newacronym
[
  longplural={Boltzmann Machines}
]
{bm}{BM}{Boltzmann Machine}
\newacronym
[
  longplural={Variational Autoencoders}
]
{vae}{VAE}{Variational Autoencoder}
\newacronym
[
  longplural={Long-Short Term Memory networks}
]
{lstm}{LSTM}{Long-Short Term Memory network}
\newacronym
[
  longplural={Probability Density Functions}
]
{pdf}{PDF}{Probability Density Function}
\newacronym
[
  longplural={Recurrent Neural Networks}
]
{rnn}{RNN}{Recurrent Neural Network}
\newacronym
[
  longplural={Generative Adversarial Networks}
]
{gan}{GAN}{Generative Adversarial Network}
\newacronym
[
  longplural={Restricted Boltzmann Machines}
]
{rbm}{RBM}{Restricted Boltzmann Machine}
\newacronym
[
  longplural={Digital Twins}
]
{dt}{DT}{Digital Twin}

\newacronym{itu}{ITU}{International Telecommunication Union}
\newacronym{etsi}{ETSI}{European Telecommunication Standards Institute}
\newacronym{api}{API}{Application Program Interface}
\newacronym{ue}{UE}{User Equipment}
\newacronym{PC}{PC}{personal computer}
\newacronym{RAN}{RAN}{radio access network}
\newacronym{CN}{CN}{core network}
\newacronym{MS}{MS}{mobile station}
\newacronym{soa}{SoA}{state of the art}

\newacronym{genai}{GenAI}{Generative AI}

\newacronym{ITU-R}{ITU-R}{International Telecommunications Union - Radiocommunication Sector}
\newacronym{IMT-Advanced}{IMT-Advanced}{International Mobile Telecommunications Advanced}
\newacronym{4G}{4G}{fourth-generation of mobile phone communications and Internet access technology}

\newacronym{rlhf}{RLHF}{Reinforcement Learning with Human Feedback}
\newacronym{3gpp}{3GPP}{Third Generation Partnership Project}
\newacronym{GSM}{GSM}{Global System for Mobile Communications}
\newacronym{UMTS}{UMTS}{Universal Mobile Telecommunications System}
\newacronym{HSPA}{HSPA}{High Speed Packet Access}
\newacronym{lte}{LTE}{Long-Term Evolution}
\newacronym{lte-a}{LTE-A}{Long-Term Evolution Advanced}

\newacronym{m2m}{M2M}{Machine-to-machine}

\newacronym{OSS}{OSS}{Operations Support System}

\newacronym{tmn}{TMN}{Telecommunications Management Network}

\newacronym{e-UTRAN}{e-UTRAN}{evolved Universal Terrestrial Radio Access Network}
\newacronym{eNB}{eNB}{e-UTRAN NodeB}
\newacronym{gNB}{gNB}{gNodeB}
\newacronym{EPC}{EPC}{Evolved Packet Core}

\newacronym{MBMS}{MBMS}{Multimedia and Broadcast Multicast Service}
\newacronym{eMBMS}{eMBMS}{Evolved MBMS}
\newacronym{SFN}{SFN}{single-frequency network}
\newacronym{MBSFN}{MBSFN}{MBMS single-frequency network}
\newacronym{BM-SC}{BM-SC}{Broadcast/Multicast Service Center}
\newacronym{MBMS GW}{MBMS GW}{MBMS Gateway}
\newacronym{MME}{MME}{Mobility Management Entity}
\newacronym{MCE}{MCE}{Multi-cell/multicast Coordinating Entity}
\newacronym{SYNC}{SYNC}{synchronization}
\newacronym{MCCH}{MCCH}{Multicast Control Channel}
\newacronym{MTCH}{MTCH}{Multicast Traffic Channel}
\newacronym{MCH}{MCH}{Multicast Channel}
\newacronym{PMCH}{PMCH}{Physical Multicast Channel}
\newacronym{PDSCH}{PDSCH}{Physical Downlink Shared Channel}
\newacronym{tmforum}{TMForum}{TeleManagement Forum}

\newacronym{sop}{SOP}{Sentence Order Prediction}
\newacronym{fid}{FID}{Fréchet Inception Distance}
\newacronym{is}{IS}{Inception Score}
\newacronym{mlm}{MLM}{Masked Language Modeling}
\newacronym{nsp}{NSP}{Next Sentence Prediction}
\newacronym{blue}{BLUE}{Bilingual Evaluation Understudy}
\newacronym{cer}{CER}{Concept Error Rate}

\newacronym{hss}{HSS}{Home Subscriber Server}

\newacronym{sft}{SFT}{Supervised Fine-Tuning}
\newacronym{gpu}{GPU}{Graphics Processing Unit}
\newacronym{IEEE}{IEEE}{Institute of Electrical and Electronics Engineers}
\newacronym{WiMAX}{WiMAX}{Worldwide Interoperability for Microwave Access}
\newacronym{ASN}{ASN}{access service network}
\newacronym{ASN-GW}{ASN-GW}{ASN gateway}
\newacronym{CSN}{CSN}{Connectivity Service Network}
\newacronym{oran}{O-RAN}{Open Radio Access Network}
\newacronym{ric}{RIC}{Radio Intelligent Controller}
\newacronym{rt}{RT}{Real-Time}
\newacronym{uav}{UAV}{Unidentified Aerial Vehicle}
\newacronym{v2x}{V2X}{Vehicle-To-Everything}

\newacronym{nwdaf}{NWDAF}{Network Data Analytics Function}
\newacronym{pcf}{PCF}{Policy Control Function}
\newacronym{mtlf}{MTLF}{Model Training Logical Function}
\newacronym{anlf}{ANLF}{Analytics Logical Function}

\newacronym{PA}{PA}{power amplifier}
\newacronym{ecgan}{ECGAN}{Enhanced Capsule Generation Adversarial Network}

\newacronym{NI}{NI}{National Instruments}

\newacronym{TDD}{TDD}{time-division duplex}
\newacronym{FDD}{FDD}{frequency-division duplex}
\newacronym{UDP}{UDP}{User Datagram Protocol}
\newacronym{APP}{APP}{application}
\newacronym{mac}{MAC}{medium access control}
\newacronym{phy}{PHY}{physical}
\newacronym{RLC}{RLC}{radio link control}
\newacronym{sdap}{SDAP}{service data adaptation protocol}
\newacronym{FIFO}{FIFO}{first-in first-out}
\newacronym{CRC}{CRC}{cyclic redundancy check}
\newacronym{SAP}{SAP}{service access point}
\newacronym{FEC}{FEC}{forward error correction}
\newacronym{IF}{IF}{intermediate frequency}
\newacronym{RF}{RF}{radio frequency}
\newacronym{mimo}{MIMO}{multiple-input and multiple-output}
\newacronym{MCS}{MCS}{modulation and coding scheme}

\newacronym{SPC}{SPC}{superposition coding}
\newacronym{SVC}{SVC}{Scalable Video Coding}
\newacronym{GM}{GM}{generic multicasting}
\newacronym{SCM}{SCM}{superposition coded multicasting}
\newacronym{SIC}{SIC}{successive interference cancellation}

\newacronym{st}{ST}{secondary transmitter}
\newacronym{pt}{PT}{primary transmitter}
\newacronym{sr}{SR}{secondary receiver}
\newacronym{pr}{PR}{primary receiver}
\newacronym{su}{SU}{secondary user}
\newacronym{pu}{PU}{primary user}

\newacronym{awgn}{AWGN}{additive white Gaussian noise}

\newacronym{cdf}{CDF}{cumulative density function}
\newacronym{ccdf}{CCDF}{complementary CDF}
\newacronym{iid}{i.i.d.}{independent and identically distributed}
\newacronym{rf}{RF}{radio frequency}
\newacronym{hbf}{HBF}{Hybrid Beamforming}

\newacronym{dd}{DD}{Device-to-Device}
\newacronym{ddu}{DDU}{Device-to-Device user}
\newacronym{dds}{DDS}{Device-to-Device system}
\newacronym{ddt}{DT}{DDU transmitter}
\newacronym{ddr}{DR}{DDU receiver}

\newacronym{bs}{BS}{base station}
\newacronym{bsu}{BSU}{base station associated user}
\newacronym{bsas}{BSAS}{base station associated system}
\newacronym{bst}{BT}{BSU transmitter}
\newacronym{bsr}{BR}{BSU receiver}

\newacronym{epg}{EPG}{energy per goodbit}
\newacronym{mepg}{MEPG}{modified energy per goodbit}
\newacronym{ee}{EE}{energy efficiency}
\newacronym{se}{SE}{spectral efficiency}

\newacronym{wrt}{w.r.t.}{with respect to}

\newacronym{kkt}{KKT}{Karush-Kuhn-Tucker}
\newacronym{al}{AL}{Active Learning}
\newacronym{admm}{ADM}{Alternating Directing Method}
\newacronym{cr}{CR}{cognitive radio}
\newacronym{ssi}{SSI}{soft-sensing information}
\newacronym{csi}{CSI}{Channel State Information}
\newacronym{qsi}{QSI}{queue state information}
\newacronym{el}{EL}{enhancement layer(s)}
\newacronym{snr}{SNR}{signal-to-noise ratio}

\newacronym{NAL}{NAL}{network abstraction layer}
\newacronym{QP}{QP}{quantization parameter}

\newacronym{ofdm}{OFDM}{orthogonal frequency-division multiplexing}
\newacronym{ofdma}{OFDMA}{orthogonal frequency-division multiple access}
\newacronym{tdma}{TDMA}{time division multiple access}

\newacronym{PUSC}{PUSC}{partial usage of the subchannels}
\newacronym{CFO}{CFO}{carrier frequency offset}
\newacronym{I/Q}{I/Q}{in-phase and quadrature-phase}
\newacronym{ASK}{ASK}{amplitude-shift keying}
\newacronym{PSK}{PSK}{phase-shift keying}
\newacronym{BPSK}{BPSK}{binary phase-shift keying}
\newacronym{QPSK}{QPSK}{quadrature phase-shift keying}
\newacronym{QAM}{QAM}{quadrature amplitude modulation}
\newacronym{PSNR}{PSNR}{peak signal-to-noise ratio}
\newacronym{PELR}{PELR}{packet error and loss rate}

\newacronym{kNN}{\textit{k}-NN}{\textit{k}-nearest neighbor algorithm}
\newacronym{SVM}{SVM}{support vector machines}
\newacronym{nn}{NN}{neural network}
\newacronym{NN}{NN}{neural network}
\newacronym{dnn}{DNN}{deep neural network}
\newacronym{RBF}{RBF}{radial basis function}
\newacronym{RMSE}{RMSE}{root mean squared error}
\newacronym{mse}{MSE}{mean squared error}
\newacronym{lmse}{LMSE}{linear mean square-error estimator}

\newacronym{R2}{$R^2$}{coefficient of determination}

\newacronym{KAUST}{KAUST}{King Abdullah University of Science and Technology}
\newacronym{GSA}{GSA}{Global mobile Suppliers Association}

\newacronym{VoD}{VoD}{video on demand}
\newacronym{HEVC}{HEVC}{High Efficiency of Video Coding}
\newacronym{DASH}{DASH}{Dynamic Adaptive Streaming over HTTP}

\newacronym{PUT}{PUT}{people using television}

\newacronym{ADTVS}{ADTVS}{Audience Driven live TV Scheduling}

\newacronym{arq}{ARQ}{automatic repeat request}

\newacronym{harq}{HARQ}{hybrid automatic repeat request}

\newacronym{sdp}{SDP}{semi-definite programming}

\newacronym{tcp}{TCP}{transmission control protocol}

\newacronym{e2e}{E2E}{end-to-end}

\newacronym{ran}{RAN}{radio access network}
\newacronym{cran}{CRAN}{cloud radio access network}
\newacronym{udcran}{UD-CRAN}{ultra-dense CRAN}
\newacronym{dran}{DRAN}{distributed radio access network}
\newacronym{hcran}{H-CRAN}{hybrid cloud radio access network}
\newacronym{hetnet}{HetNet}{heterogeneous network}
\newacronym{vcran}{V-CRAN}{virtualized CRAN}
\newacronym{ecran}{E-CRAN}{edge-CRAN}
\newacronym{hvcran}{H-VCRAN}{hybrid-virtualized CRAN}

\newacronym{bbu}{BBU}{baseband processing unit}
\newacronym{rrh}{RRH}{remote radio head}
\newacronym{ru}{RU}{radio unit}
\newacronym{rs}{RS}{remote site}
\newacronym{cs}{CS}{central site}

\newacronym{rru}{RRU}{radio resource unit}
\newacronym{rb}{RB}{resource block}
\newacronym{hpn}{HPN}{high-power node}
\newacronym{lpn}{LPN}{low-power node}
\newacronym{mabs}{MaBS}{macro basestation}

\newacronym{comp}{CoMP}{coordinated multi-point}
\newacronym{ranaas}{RANaaS}{RAN-as-a-Service}

\newacronym{rof}{RoF}{radio over fiber}
\newacronym{wdm}{WDM}{Wavelength Division Multiplexing}
\newacronym{dls}{DLS}{distributed large scale}

\newacronym{qos}{QoS}{quality of service}
\newacronym{qoe}{QoE}{quality of experience}
\newacronym{qee}{QEE}{quality of energy-efficiency}

\newacronym{gg}{GG}{group-to-group}
\newacronym{ht}{HT}{hyper-transceiver}

\newacronym{fh}{FH}{fronthaul}
\newacronym{dl}{DL}{downlink}
\newacronym{ul}{UL}{uplink}

\newacronym{cp}{CP}{Cell-Processing}
\newacronym{up}{UP}{User-Processing}

\newacronym{sla}{SLA}{Service-Level Agreement}

\newacronym{kr}{KR}{knowledge representation}

\newacronym{co}{CO}{center office}

\newacronym{krr}{KRR}{Knowledge Representation and Reasoning}
\newacronym{du}{DU}{digital unit}
\newacronym{lc}{LC}{Line-Card}

\newacronym{onu}{ONU}{optical network unit}
\newacronym{olt}{OLT}{optical line terminal}
\newacronym{osw}{OSW}{optical switch}

\newacronym{es}{ES}{ethernet switch}

\newacronym{ppp}{PPP}{Poisson point process}

\newacronym{mppp}{MPPP}{marked Poisson point process}

\newacronym{sinr}{SINR}{signal to noise and interference ratio}

\newacronym{sir}{SIR}{signal to interference ratio}

\newacronym{mbs}{MBS}{macro basestation}
\newacronym{ap}{AP}{access point}
\newacronym{fap}{FAP}{femto-cell access point}
\newacronym{sap}{SAP}{small-cell access point}
\newacronym{iot}{IoT}{Internet of Things}
\newacronym{ti}{TI}{Tactile Internet}
\newacronym{ntn}{NTN}{Non-Terrestrial Network}
\newacronym{lsm}{LSM}{linear scalarizing method}

\newacronym{lp}{LP}{Low-Priority}
\newacronym{hp}{HP}{High-Priority}
\newacronym{lpu}{LPU}{Low-Priority user}
\newacronym{hpu}{HPU}{High-Priority user}
\newacronym{lps}{LPS}{Low-Priority system}
\newacronym{hps}{HPS}{High-Priority system}

\newacronym{ietf}{IETF}{Internet Engineering Task Force}

\newacronym{ttm}{TTM}{time to market}
\newacronym{udn}{UDN}{ultra-dense network}

\newacronym{capex}{CAPEX}{capital expenditure}
\newacronym{opex}{OPEX}{operational expenditure}

\newacronym{cpri}{CPRI}{common public radio interface}
\newacronym{otn}{OTN}{optical transport network}
\newacronym{pon}{PON}{passive optical network}
\newacronym{twdm}{TWDM}{time and wavelength division multiplexing}

\newacronym{ec}{EC}{Edge-Cloud}
\newacronym{cc}{CC}{Central-Cloud}

\newacronym{mmw}{m-Wave}{Milli-Meter wave}

\newacronym{gops}{GOPS}{giga operation per second}
\newacronym{mops}{MOPS}{mega operation per second}

\newacronym{ipr}{IP}{Intellectual Property}

\newacronym{ip}{IP}{internet protocol}
\newacronym{rlc}{RLC}{radio link control}
\newacronym{pdcp}{PDCP}{packet data convergence protocol}

\newacronym{mno}{MNO}{mobile network operator}
\newacronym{prb}{PRB}{physical resource block}
\newacronym{mi}{MI}{modulation index}

\newacronym{wifi}{WiFi}{wireless local area network}
\newacronym{cpu}{CPU}{central processing unit}
\newacronym{vcpu}{VCPU}{virtual CPU}
\newacronym{vm}{VM}{virtual machine}

\newacronym{urs}{UrS}{user requested service}

\newacronym{npcomplete}{NP-complete}{nondeterministic polynomial-time complete}

\newacronym{rsf}{RSF}{radio sub-frame}
\newacronym{siso}{SISO}{single-input single-output}
\newacronym{ram}{RAM}{random access memory}
\newacronym{xr}{XR}{Extended Reality}

\newacronym{nef}{NEF}{Network Exposure Function}

\newacronym{vr}{VR}{Virtual Reality}
\newacronym{agv}{AGV}{Automated Guided Vehicle}

\newacronym{rssi}{RSSI}{Received Signal Strength Indicator}

\newacronym{mec}{MEC}{mobile edge computing}
\newacronym{co2}{CO$_{2}$}{carbo dioxide}

\newacronym{cfp}{CFP}{communication function processing}
\newacronym{ptp}{PTP}{precision time protocol}

\newacronym{mdt}{MDT}{Model Drive Test}

\newacronym{voip}{VoIP}{voice over Internet protocol}
\newacronym{sdn}{SDN}{Software Defined Network}

\newacronym{da}{DA}{data analytics}

\newacronym{kpi}{KPI}{key performance indicator}

\newacronym{noc}{NOC}{Network Operations Centre}
\newacronym{fso}{FSO}{Field Service Operations}

\newacronym{fsmc}{FSMC}{finite state markov chain}

\newacronym{nr}{NR}{new radio}
\newacronym{gnbcu}{gNB-CU}{gNB central unit}
\newacronym{gnbdu}{gNB-DU}{gNB distributed unit}
\newacronym{ecpri}{eCPRI}{common public radio interface}
\newacronym{fl}{FL}{federated learning}
\newacronym{rsrq}{RSRQ}{Reference Signal Received Quality}
\newacronym{rsrp}{RSRP}{Reference Signal Received Power}
\newacronym{urllc}{URLLC}{ultra-reliable low-latency communications}
\newacronym{embb}{eMBB}{enhanced mobile broadband}

\newacronym{mae}{MAE}{modified autoencoder}
\newacronym{mtc}{MTC}{machine type communication}
\newacronym{mmtc}{mMTC}{massive machine type communication}
\newacronym{pca}{PCA}{principal component analysis}

\newacronym{cps}{CPS}{cyber-physical system}
\newacronym{gnb}{gNB}{gNodeB}
\newacronym{ref}{REF}{reliability enhancement feature}
\newacronym{nfo}{NFO}{network level feature orchestrator}
\newacronym{dc}{DC}{data center}
\newacronym{vnf}{VNF}{virtual network function}
\newacronym{nssmf}{NSSMF}{Network Slice Subnet Management Function}
\newacronym{ai}{AI}{Artificial Intelligence}
\newacronym{rl}{RL}{Reinforcement Learning}
\newacronym{ddpg}{DDPG}{deep deterministic policy gradient}
\newacronym{dqn}{DQN}{deep Q-networks}
\newacronym{sac}{SAC}{soft actor-critic}
\newacronym{a2c}{A2C}{advantage actor-critic}
\newacronym{td3}{TD3}{twin delayed deep deterministic policy gradient algorithm}
\newacronym{poc}{PoC}{proof of concept}
\newacronym{cvae}{CVAE}{Conditional Variational AutoEncoder}
\newacronym{oam}{OAM}{Operation, Administration, and Maintenance}

\newcommand*\myglsentry[1]{%
  \protect\ifglsused{#1}{%
    \glsentryshort{#1}%
  }{%
    \glsentrylong{#1}%
  }%
}

\newacronym{v2n}{V2N}{Vehicle To Network}
\newacronym{v2n2v}{V2N2V}{Vehicle to Network to Vehicle}
\newacronym{fdd}{FDD}{Frequency Division Duplexing}
\newacronym{cots}{COTS}{Commercial Off-The-Shelf}
\newacronym{dme}{DME}{Dedicated Measurement Equipment}
\newacronym{cgan}{CGAN}{Conditional Generative Adversarial Network}
\newacronym{dtpc}{DT-PC}{Digital Twin for Protocol and Connectivity}
\newacronym{jcas}{JCAS}{joint communication and sensing}
\newacronym{aql}{AQL}{Action-conditioned Q-Learning}

%% file: sections/section1_introduction.tex
\section{Introduction}

\par{The deployment of \gls{5g} has seen a significant application of \gls{ai} technologies to enhance efficiency, reliability, and user experience. Such technologies include both \gls{3gpp}-defined use cases such as energy-saving, load balancing and mobility optimization \cite{3gpp.38.843}, as well as independent implementations of network functionality, such as resource allocation, network optimization, and anomaly detection \cite{8539642}. These technologies generally involve ``discriminative'' approaches, wherein \gls{ml}-trained models classify input data into predefined categories, by learning the boundaries between classes. While discriminative models have shown their utility in automating certain aspects of mobile network functionality, they are also restricted in their capacity for decision-making when confronted with complex problems that demand critical thinking, i.e., analysis and evaluation of information in a systematic and logical manner. In such instances, advanced techniques, such as reasoning and planning, are necessary.}

\par{Reasoning is a critical capability for AI systems, enabling them to interpret, infer, and make decisions based on the information they process. A recent survey highlighted several use-cases in mobile networks that can benefit from reasoning, including use cases using expert systems, such as \gls{qa}, intent-based networking automation, and customer assistance \cite{telecom5010006}.}

\par{While reasoning draws conclusions based on logical analysis of information, AI planning algorithms create sequences of actions to achieve specific goals, focusing on procedural aspect of decision making.}

\par{Symbolic \gls{ai} approaches were traditionally used to solve reasoning and planning problems. In symbolic \gls{ai}, knowledge is represented explicitly in symbolic form that is understandable both by computers and humans. This allows symbolic systems to operate with clear, well-defined rules and facts making them well-suited for logic tasks such as reasoning. However such approaches may suffer from scalability issues, as the cost of acquiring and formalizing the required knowledge compounds. They are also brittle, meaning that they are intolerant to noisy data that can occur during knowledge acquisition \cite{Cummings_2021}. 
Moreover, the computational complexity of generic reasoning tasks is usually high. For example, satisfiability of boolean formula which is a very common symbolic model for many reasoning tasks is \gls{npcomplete}, meaning there is little chance of finding efficient solutions.
Recently, neuro-symbolic approaches combining discriminative and symbolic \gls{ai} 
have emerged as an attempt to address complex problems that require reasoning\cite{YU2023105}.

\par{In mobile networks, \gls{genai} technologies are beginning to complement the established use of discriminative \gls{ai}, marking a significant advancement in network capabilities and services. Unlike their discriminative counterparts, \gls{genai} models create new data, such as network configurations, mobile subscription plans, provide \gls{qa}-type of expert systems, generate code, etc. \cite{Karapantelakis2024-zg, thanos_alex_ICML}. Specifically, \gls{genai} models such as \glspl{llm} have been empirically observed to exhibit an emergence of reasoning-like abilities, even if it is not yet clearly understood whether these models can in fact reason \cite{huang2023reasoning}.}

\par{The contribution of this study is threefold. First, it reviews the \gls{soa} of \gls{genai} algorithms with critical thinking capabilities. These algorithms are categorized based on the nature of the problems they solve. Second, it describes areas of mobile networks in form of potential use-cases where critical thinking-based algorithms can be effectively deployed. 
Finally and with the intention to inspire future research, it correlates the GenAI algorithms with the telecom use cases.}

\par{This paper is structured as follows: 
Section \ref{section2_background} outlines the background information on the evolution of telecommunication networks, \gls{genai}, and symbolic knowledge representation and reasoning, necessary to help readers understand the subject matter of the paper.
In section \ref{section3_relevantApproaches}, we introduce a classification system for critical thinking  methodologies and \gls{genai} approaches to solve critical thinking tasks. 
Section \ref{section4_telecom_use_cases} describes use-cases that involve critical thinking i.e. reasoning and planning, in telecom networks. It also maps the critical thinking based tasks to the relevant GenAI based solution methods.
GenAI based  methods face many challenges in the large scale, real-life implementations. These challenges and associated research opportunities are discussed in Section \ref{section5_challenges_opportunities}.
Finally, section \ref{section6_conclusion} summarizes the contributions and identifies potential directions for future research.}

%% file: sections/section2_background.tex
\section{Background}
\label{section2_background}

\par{In this section, we provide an overview of the historical progression of mobile networks, and realize that they have reached a level of complexity where \gls{ai} becomes indispensable for cost-effective network lifecycle management. Next, we examine the current \gls{soa} in \gls{genai}, focusing specifically on the rapid progress within four of the most promising categories of contemporary algorithms. These advancements are poised to address the complexity issues highlighted earlier. Lastly, we outline the field of Knowledge representation and reasoning which is the foundation on which critical thinking - reasoning and planning - techniques are built.}


\subsection{Evolution of telecom networks}
\label{subsec:network_evolution}

\begin{figure}[t!]
\centerline{\includegraphics[width=0.5\textwidth]{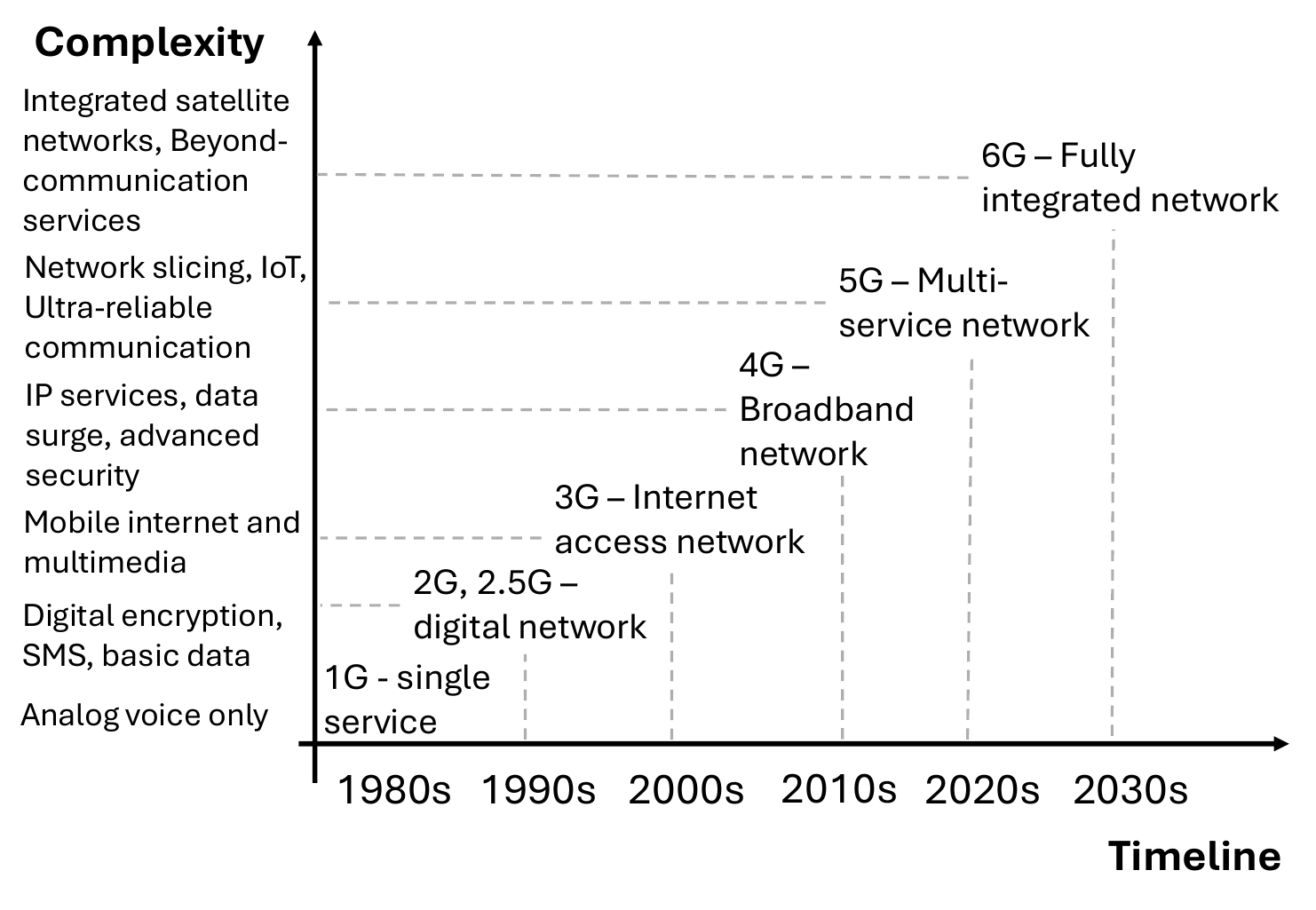}}
\caption{Increasing complexity of network-related operations over different generations of mobile networks.}
\label{fig1_complexity}
\end{figure}

Figure \ref{fig1_complexity} illustrates the growing complexity in supporting mobile network operation across generations of mobile networks. 

\par{Starting with the \gls{1g}, which introduced mobile analog voice across multiple standards, the industry transitioned to the second generation \gls{2g}. \gls{2g} brought digital voice and basic messaging services to the forefront, setting a global interoperability standard. The interim 2.5G introduced packet-switched mobile data, laying the groundwork for the \gls{3g}, which aimed to bridge mobile networks with the global internet. This effort was further advanced by the \gls{4g}, which embraced internet standards, and the fifth generation \gls{5g}, which incorporated cloud computing principles into its foundation. Both \gls{4g} and \gls{5g} have led to a significant increase in the number of connected devices, spanning both personal and industrial uses, particularly with the Internet of Things (IoT). The \gls{6g} is anticipated to extend these developments, integrating more comprehensive cloud computing principles and artificial intelligence to support \gls{ti} and \glspl{ntn}.}

\par{The evolution of mobile network management from \gls{1g} to \gls{6g} highlights a significant increase in complexity across both operations and planning activities. In the early days of \gls{1g} and \gls{2g}, network management was straightforward enough to proceed without any artificial intelligence (\gls{ai}), focusing primarily on voice communication with relatively simple infrastructure. The transition to \gls{3g} introduced more complexity, necessitating simple automation techniques, such as scripts or rule-based systems, to cope with the added demands of data services. However, with the advent of \gls{4g}, the emphasis on data-centric communication called for applying discriminative \gls{ai} techniques, including pattern recognition and prediction, to enhance efficiency and manage the growing data traffic. As we move into the era of \gls{5g} and look towards \gls{6g}, the network landscape becomes even more complex, requiring sophisticated \gls{ai} reasoning techniques, as monitoring and management by human operators alone is becoming increasingly impractical. This complexity is driven by the need to support massive \gls{iot} connectivity, ultra-reliable low-latency communications, and extremely high data rates, demanding \gls{ai} systems capable of advanced decision-making, predictive analytics, and real-time adaptation to ensure seamless network performance and reliability.}



\subsection{Generative Modeling}
\label{subsection_sec2_generative_modeling}

\gls{genai} encompasses a broad range of applications aimed at producing novel content, ideas, or solutions. It predominantly relies on generative modeling, a specific set of techniques that train models to understand and replicate the complex distributions of data they learn from, such as images, audio, and text. 

Generative modeling techniques have evolved significantly over time, starting with the development of Markov Chains in the 1950s, which are statistical models used to predict the likelihood of sequential events. The introduction of neural network-based \glspl{bm} in the 1980s marked a significant milestone, making generative models more applicable to real-world problems \cite{6302930}.  In the mid-2000s, \glspl{rbm} advanced generative models in fields such as \gls{nlp} and computer vision, with their stacking forming \glspl{dbn} displaying improved performance.

Since 2014, the landscape of \gls{genai} has witnessed a transformative shift, largely driven by improvements in computational infrastructure and the evolution of generative modeling algorithms. We highlight the four most distinguished families of algorithms:
\begin{itemize}

\item \glspl{gan} operate through a system involving two neural network models: a Generator and a Discriminator \cite{NIPS2014_5ca3e9b1}. The training of these models constitutes a minimax game, wherein the Generator aims to reduce the instances in which the Discriminator identifies its produced samples as fake.

\item \glspl{vae} encode input data into a compressed, latent representation, then reconstruct the input from this representation \cite{Kingma2014}. They differ from traditional autoencoders by introducing a probabilistic approach to the encoding process, allowing them to generate new data points similar to the input data. 

\item Transformers marked a significant departure from previous sequence modeling approaches \cite{vaswani2017attention}. Unlike models that rely on recurrent or convolutional layers that serially process input data, transformers use self-attention mechanisms to process input data in parallel, significantly improving efficiency and performance on tasks requiring an understanding of long-range dependencies in data. The transformer architecture enabled  creation of \glspl{llm} \cite{zhao_survey_2023, sun_survey_2024}. In the current \gls{soa} research, three advancements stand out, enabling development of \glspl{llm} with enhanced critical thinking capabilities. First, support for different modalities in addition to natural language, including image and audio \cite{yin2023survey}. Second, the application of the Mixture of Experts (MoE) approach, particularly in Mixtral model and its derivatives, has further enhanced the performance of \glspl{llm} by utilizing a network of experts to process inputs, while reducing their computational footprint \cite{jiang2024mixtral}. Third, the increase in context length, i.e., the number of tokens, such as words, that an \gls{llm} can take as input when generating responses. Context is a crucial characteristic of \glspl{llm}, with newer models continuously increasing their capacity to parse longer contexts, thereby enabling more complex tasks and richer outputs.

\item Finally, diffusion models work by gradually adding noise to data over a series of steps to create a distribution and then learning to reverse this process to generate data from noise \cite{DBLP:journals/corr/abs-2006-11239}. Introduced around the early 2020s, these models stand out for their effectiveness in generating detailed and diverse outputs, rivaling and sometimes surpassing the capabilities of GANs. 
\end{itemize}

\begin{table}[t!]
\caption{Taxonomy of Generative Models}
\begin{tabular}{|p{0.2\columnwidth}|p{0.2\columnwidth}|p{0.45\columnwidth}|}
\hline
\textbf{Model Type} & \textbf{Category} & \textbf{Applications} \\ \hline
Transformers & Self-attention-based Models & \gls{nlp} (text completion, translation, summarization, etc.) \cite{LIU2023100017} \\ \hline
\glspl{gan} & Implicit Generative Models & Image, audio and video generation \cite{jabbar2020survey} \\ \hline
\glspl{vae} & Explicit Generative Models & Data generation/augmentation \cite{chadebec2021data}, anomaly detection \cite{8999265} \\ \hline
Diffusion Models & Stochastic Generative Models & Computer Vision (image and video generation, image processing, semantic segmentation) \cite{yang2024diffusion} \\ \hline
\end{tabular}
\label{table:taxonomy_applications}
\end{table}

Table \ref{table:taxonomy_applications} shows categories of generative models\footnote{In the context of this document, the terms \textit{generative models} and \textit{generative modeling} are used interchangeably to denote algorithms used to train \gls{ml} models for content generation.}. The ``category'' column refers to the foundational approach each family of algorithms uses to train a generative model. Transformers use attention mechanisms to prioritize relevant parts of the input data. \glspl{gan} generate realistic data outputs by imitating the data distribution without defining it explicitly. \glspl{vae} directly model and learn the data's probability distribution to generate new samples. Finally, diffusion models employ stochastic processes to transform noise into data resembling the training set through a gradual, reverse procedure.

\subsection{Knowledge Representation and Reasoning (KRR)}

Critical thinking plays a pivotal role in enabling \gls{ai} systems to effectively process and utilize knowledge for reasoning, and making well-informed decisions. The field dedicated to storing and accessing information in a manner conducive to this cognitive process is referred to as \gls{krr}. Symbolic representations have traditionally served as the primary means for encoding, storing, and managing knowledge. This section provides some background information with regards to how contemporary methodologies also incorporate embedded forms and facilitate the manipulation of knowledge by neural networks, including \gls{genai} techniques.



\begin{table}[t!]
\caption{Comparison of Knowledge Representation Formalisms}
\label{tab:kr_formalisms_revised}
\centering
\begin{tabular}{|p{0.3\columnwidth}|p{0.6\columnwidth}|}
\hline
\textbf{\gls{krr} type} & \textbf{Types of Algorithms Used for Reasoning} \\ \hline
Semantic Networks, Ontologies & Graph algorithms, Path-finding algorithms \\ \hline
Frames & Inheritance algorithms, Slot-filling algorithms \\ \hline
Production Rules & Rule-based engines, Forward/Backward chaining reasoning \\ \hline
Formal Logic & Resolution, Automated theorem proving, Logic programming \\ \hline
Probabilistic Logic & Bayesian networks, Markov models, Fuzzy logic \\ \hline
Model-based & Deep learning models (e.g., transformers, GNNs), Belief networks, Embeddings, etc. \\ \hline
\end{tabular}
\end{table}

Table \ref{tab:kr_formalisms_revised} compares traditional and modern formalisms for representing knowledge, highlighting the evolution of techniques from structured, rule-based systems to advanced, AI-driven models. Traditional knowledge representation methods are Semantic Networks, Frames, Production Rules, Formal and Probabilistic Logics \cite{PATEL2018542}. These rely on explicit methodologies like graph algorithms for Semantic Networks, which facilitate associative reasoning by connecting concepts and relationships in a network-like structure. Frames use inheritance and slot-filling algorithms to support hierarchical reasoning, organizing knowledge into structured entities with attributes and values. Production Rules apply rule-based engines and logic for conditional reasoning, operating on a set of if-then rules to derive conclusions. In Formal Logic, resolution techniques are used to infer new facts from known truths. Representations based on Probabilistic Logic model ambiguous situations and make conclusions under conditions of incomplete or uncertain information. 

With the success of discriminative and generative \gls{ai}, model-based formalisms for representing knowledge have emerged \cite{koudouridis_evaluation_2024}. The evolution of \gls{krr} from traditional to modern formalisms reflects a broader shift in the field of \gls{ai}, from rule-based to data-driven approaches, enabling more nuanced, flexible, and powerful systems capable of addressing the complexities of real-world data and applications.

%% file: sections/section3_relevant_approaches.tex
\section{\gls{genai} and critical thinking}
\label{section3_relevantApproaches}

\begin{table}[t!]
\caption{Types of reasoning, with examples}
\centering
\begin{tabular}{|p{0.45\columnwidth}|p{0.45\columnwidth}|}
\hline
\textbf{Inductive reasoning} & \textbf{Deductive reasoning}  \\
\hline
\textit{Fact 1}: In rural region A, an upgrade of radio units from model X to model Y led to 20\% power savings.
\newline
\textit{Fact 2}: In urban region B, a similar upgrade from model X to Y provided 40\% power savings.
\newline
\newline
\textbf{Rule (conclusion)}:\newline
Upgrade of units model X to model Y results with power savings.
& 
\textit{Fact 1}: Vendor A's radio units offer up to 20\% power saving compared to vendor B's units.
\newline
\textit{Fact 2}: Operator O's RAN comprises of 60\% of radio units from vendor B which are going to be replaced in the next year.
\newline
\newline
\textbf{Fact (conclusion)}: If the operator replaces all units of vendor A the power consumption will be reduced by up to 12\% in the next year.
\newline
\\
\hline

\textbf{Analogical reasoning} & \textbf{Abductive reasoning}  \\
\hline
\textit{Fact 1}: Migration from 3G to 4G led to 10\% experienced increase in average revenue per user (ARPU) for operator O.
\newline
\textit{Fact 2}: Improved network technology leads to increased revenue.
\newline
\textit{Fact 3}: Operator P needs to increase ARPU.
\newline
\newline
\textbf{Fact (conclusion)}:\newline
Operator P can consider migrating from 3G to 4G in order to increase ARPU. 
&
\textit{Fact 1}: Customer C is experiencing reduced broadband throughput.
\newline
\textit{Rule 1}: When a network cell is congested, the connected customers experience temporarily reduced throughput.
\newline
\textit{Rule 2}: When two or more network cells interfere, the connected customers experience reduced throughput.
\newline
\newline
\textbf{Fact (conclusion)}: Customer C is connected to a congested cell. \\
\hline
\end{tabular}
\label{table:reasoning_examples}
\end{table}

In this section, we establish a classification system for reasoning methods applied across the four types of \gls{genai} algorithms highlighted in section \ref{subsection_sec2_generative_modeling}. This system evaluates on two fronts: the nature of reasoning each algorithm employs and the particular reasoning task or tasks, for which each model is designed and has exhibited superior performance. Before presenting the taxonomy, we start with a succinct introduction to the different reasoning approaches and their corresponding tasks. 

\subsection{Reasoning Types and Tasks}
\label{subsec:rtypes_tasks}

Four forms of reasoning, or reasoning ``types'', that are often referenced in the literature are deduction, abduction, analogy and induction \cite{Cellucci1998-CELTSO}. Deduction involves deriving specific conclusions from general premises, while abduction pertains to generating hypotheses that explain observed phenomena. Analogy relies on identifying similarities between different entities to infer new insights, and induction involves generalizing from specific instances to broader cases. An example of these four forms using simple telecom concepts is shown in table \ref{table:reasoning_examples}.

\begin{table}[t!]
\caption{Overview of Datasets for Reasoning Problems}
\centering
\begin{tabular}{|p{2cm}|p{3cm}|p{2.5cm}|}
\hline
\textbf{Reasoning Task} & \textbf{Description}                                                                                     & \textbf{Dataset}        \\ \hline
Mathematics: \newline Algebra, \newline Geometry & Solving numerical, geometrical and algebraic problems as well as simple physics. \tablefootnote{The current focus is on the areas of mathematics that attract the widest audience, not necessarily excluding other areas.}                   & GSM8K \cite{DBLP:journals/corr/abs-2110-14168}, SVAMP \cite{patel-etal-2021-nlp}, MAWPS \cite{koncel:naacl16}, MathQA \cite{DBLP:journals/corr/abs-1905-13319}, IMO-AG-30 \cite{Trinh2024}, MATH Dataset \cite{hendrycksmath2021}                  \\ \hline
Commonsense and Language understanding           & Understanding and inferring everyday knowledge, manipulating terms according to arithmetical, logical and commonsense rules                                                            & CommonsenseQA \cite{talmor-etal-2019-commonsenseqa}, StrategyQA \cite{10.1162/tacl_a_00370}, SayCan \cite{saycan2022arxiv},  CLEVR\cite{johnson2016clevr}, CLEVRER \cite{yi2020clevrer}, Last Letters \cite{wei_chain_of_thought_2022}, Coin flip \cite{wang2023boosting}        \\ \hline
Logic/Symbolic                 & Applying logical reasoning and inference, such as deduction, induction, and abduction                                   & LogiQA 2.0 \cite{10174688} SNLI \cite{snli:emnlp2015}, MultiNLI \cite{N18-1101}          \\ \hline
\end{tabular}

\label{table:reasoning_tasks}
\end{table}

Similar to types of reasoning, we also consider reasoning tasks as \textit{cognitive activities in different domains}. Sun et al distinguish between several domains, including mathematics, logical, visual, commonsense, and others\cite{sun_survey_2024}. For various task categories, distinct datasets are available comprising problems that necessitate the use of single or multiple types of reasoning in order to solve them. Beyond pinpointing possible application areas for reasoning algorithms, these datasets play a crucial role in assessing the performance of algorithms. An abridged summary of reasoning tasks along with the datasets referenced in the literature is depicted in table \ref{table:reasoning_tasks}.

We consider three main categories of reasoning tasks. First, mathematical tasks that involve solving different types of mathematical problems, such as numerical, geometrical and algebraic. Second, commonsense reasoning tasks describe the capability to make assumptions about the nature and characteristics of typical scenarios that humans deal with on a daily basis. Such commonsense tasks may take the form of \gls{qa} pairs, for example using an open-ended, multiple choice or visually-grounded format. Other types of commonsense tasks include constrained \gls{nlg}, wherein reasoning algorithms are instructed to produce text that adheres to specific predefined rules or limitations and reading comprehension, where commonsense knowledge is used together with given text to provide correct answers. A variation of this are language understanding tasks with specifications expressed as simple logical rules, based on which generative models manipulate symbols to reach answers. Datasets include games such as ``last letter'' and ``coin flip'' \cite{wang2023boosting}. The former asks the model to concatenate the last letters of words in a name, whereas the latter asks the model to answer whether a coin is still heads up after people either flip or don’t flip the coin. Finally, more complex logical task datasets check for textual entailment, wherein one piece of text (hypothesis), logically follows from another text (premise). 

\subsection{Reasoning Approaches in \gls{genai}}

In this section, we examine the latest advancements in utilizing \gls{genai} algorithms for reasoning tasks. Our analysis employs a dual-axis classification: the first axis categorizes the reasoning tasks targeted by the approach, and the second axis identifies the reasoning type, as detailed in section \ref{subsec:rtypes_tasks}. To streamline our review of the state of the art, we divide our discussion into subsections. Each subsection addresses a distinct category of \gls{genai} algorithms, namely, transformer, GAN, CVAE and diffusion model, introduced in Table \ref{table:taxonomy_applications}.

\subsubsection{\glspl{llm}}

\begin{figure}[t!]
    \centering
    \includegraphics[width=1\columnwidth]{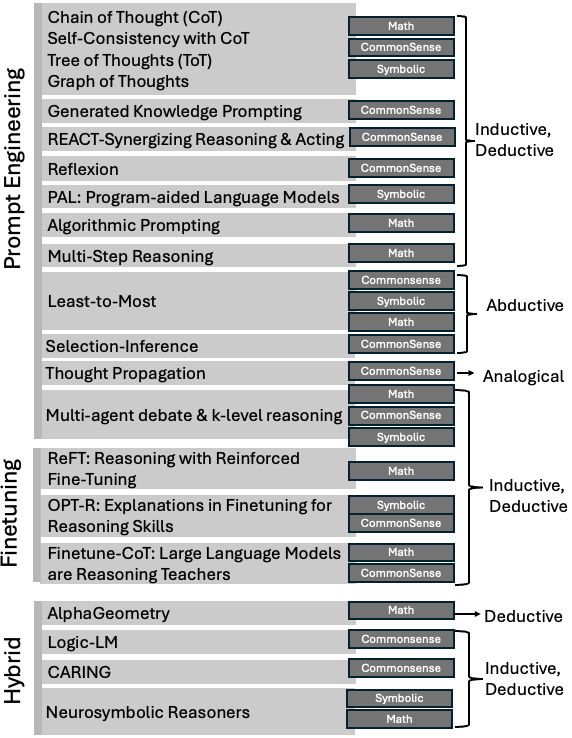}
   \caption{Reasoning approaches in \glspl{llm}}
    \label{fig:llm_taxonomy}
\end{figure}

A thorough review of reasoning approaches using \glspl{llm} was presented in \cite{huang2023reasoning}. In this section, we augment the insights of this survey by incorporating recent research that has emerged. An abbreviated version of reviewed literature is illustrated in figure \ref{fig:llm_taxonomy}.

As illustrated in the figure, we distinguish between three types of approaches, namely ``prompt engineering'', ``finetuning'' and ``hybrid''. 

Prompt engineering involves carefully crafting input queries to guide the model's responses and achieve desired outputs. It requires human input to design prompts that elicit the desired information or behavior from the model. 

On the other hand, fine-tuning is a process where pre-trained models are further trained on specific tasks or domains using labeled data. It involves adjusting model parameters to optimize performance for a particular task, making it more tailored and specialized. Both techniques play crucial roles in optimizing models for specific applications. Finally, hybrid solutions incorporate \glspl{llm} with other types of \gls{ai} algorithms, such as symbolic and discriminative models. The intention is to combine the content generation and associative memory capabilities of \glspl{llm}, with the deterministic output and analytical abilities of symbolic solvers or classifiers. \\ \\
\textit{i. Prompt Engineering Approaches}
\\ \\
A prominent family of prompt-engineering approaches relies in prompting the \gls{llm} with so-called ``thoughts'', i.e., intermediate cognitive processes or steps involved in generating responses to prompts or queries. In this study, we distinguish some notable works that incrementally increase in complexity but also perform better.
\begin{itemize}
\item The original \gls{cot} approach represents one of the pioneering strategies in prompt engineering aimed at enhancing the reasoning capabilities of \glspl{llm} \cite{wei_chain_of_thought_2022}. This method entails crafting a sequence of intermediate reasoning steps, referred to as a ``chain of thought'', to guide the \gls{llm} in tackling intricate reasoning tasks. Chain of thought prompting offers several appealing attributes. Notably, it enables models to break down multi-step problems into intermediate stages, facilitating the allocation of additional computation to tasks requiring more reasoning steps. Additionally, it enhances explainability, facilitating debugging and potential model improvement through fine-tuning. Furthermore, it operates with a few-shot mechanism, meaning that only a small number of exemplars are required to prompt \gls{cot} responses. The study compares \gls{cot} \glspl{llm} with vanilla-prompted \glspl{llm} using the GSM8K math benchmark, revealing that the \gls{cot} achieves state-of-the-art accuracy with just eight exemplars, surpassing even a finetuned GPT-3 model (see also table \ref{table:reasoning_tasks}). This superior performance extends to symbolic reasoning and commonsense tasks, showcasing the broad effectiveness of the \gls{cot} prompting method.

\item Self-consistency~\cite{wang_selfconsistency_2022} is an approach to improve on \gls{cot}, which works by sampling multiple independent \glspl{cot} and selecting the most frequent one as the final output. This proposal explores the stochasticity of the inference process, in which multiple reasoning paths can be generated towards the same goal, but some may diverge. Self-consistency assumes that reliable output is more likely to be produced. Results showed that self-consistency improves the arithmetic reasoning performance over \gls{cot}.

\item The concept of \gls{tot} draws inspiration from planning processes developed in the 1950s, wherein problem-solving involves searching through a combinatorial problem space, depicted as a tree \cite{yao_tree_2023}. It is modeled after human problem-solving abilities, reflecting the way human solvers might backtrack to previous steps if a derivation is incorrect or if they encounter an impasse, unable to make further progress toward reaching a final answer. In this context, when presented with a user query, a prompter agent engages in a multi-round conversation with the \gls{llm}, guiding the search for solutions.

\item \gls{got} represents information generated by an LLM as a flexible graph, with thoughts serving as vertices and edges denoting dependencies between them \cite{besta2024graph}. \gls{got} outperforms other prompting schemes, for example ensuring 62\% increase in the quality of sorting over \gls{tot}, while simultaneously reducing costs by more than 31\%. 
\end{itemize}

In general, \gls{cot} and its derivatives perform well in inductive and deductive reasoning tasks, and in Math, commonsense and symbolic problems.

Another approach focusing on commonsense reasoning, considers generated knowledge prompting \cite{DBLP:journals/corr/abs-2110-08387}. In this idea, two \glspl{llm} are used sequentially. The first is prompted with a few task-specific, human-curated \gls{qa} exemplars to generate knowledge statements. The second \gls{llm} is used to make predictions for each knowledge statements generated by the first, then selecting the prediction with the highest confidence. The authors found that this approach performs well with multiple-choice commonsense \gls{qa} datasets.

Another study considers grounding thoughts using external knowledge, as opposed to \gls{cot}-style of reasoning, wherein the \gls{llm} uses only its own representation to generate thoughts. The latter can lead to hallucinations, meaning that \gls{llm} may generate responses that is either factually incorrect, nonsensical, or disconnected from the input prompt. To address this issue, \gls{react} interleaves \gls{llm}-generated thought with actions taken by the \gls{llm} \cite{yao2023react}. The observation of the results from an action informs the next thought. An example of an action could be for example searching for information on the Internet, and using the returned information as knowledge for generating the next thought. \gls{react} is found to perform better than baseline approaches in commonsense \gls{qa} tasks.

In a similar fashion, Reflexion is a strategy~\cite{reflexion_2023} inspired by \gls{rl}. It aims at improving the performance of \gls{llm}-based agents and consists of three interacting components: an Actor that generates text and actions, an Evaluator that scores previously generated trajectories, and a Self-reflection module that translates the reward signal into richer verbal feedback that is reported back as input to the Actor. The method converges over multiple steps, significantly improving against \gls{cot} and \gls{react} baselines in decision-making, reasoning, and programming tasks.

In steps of the \gls{react} approach, \gls{pal} suggests thoughts to be programs, that can be offloaded to a runtime such as a Python interpreter \cite{gao2023pal}. \gls{pal} has shown to outperform \gls{cot} in symbolic and algorithmic datasets. 

Another group of approaches aims to increase reasoning capability of \glspl{llm} by introducing ways to refine the exemplars for various classes of problems. This is in contrast to \gls{cot} and its variants which used a set of human-crafted exemplars that prompted the \gls{llm}. This group of approaches is known as ``rationale refinement''.

One approach involves utilizing the algorithmic prompting method for mathematical tasks \cite{zhou2022teaching}. This method involves breaking down complex problems into smaller tasks with clear and predictable outcomes. Initially, basic algorithms such as  addition and subtraction are used to prompt the \gls{llm}. Subsequently, the \gls{llm} is presented with a series of prompts that combine these taught skills -- for example multiplication is taught using the addition skill as basis. By gradually building upon these basic skills, authors show that it is possible to create more intricate problem-solving abilities within the \gls{llm}. The rationale behind this approach is that by first teaching the \gls{llm} fundamental skills, it avoids attempting to independently derive them later, resulting in improved performance from \gls{cot} in specific mathematical tasks.

An alternative method in the rationale refinement category of approaches uses complexity-based prompting for multi-step reasoning. According to this approach, the initial thoughts within the exemplars featured in the \gls{cot} approaches are broken down into increasingly detailed thoughts \cite{fu2023complexitybased}. The authors observed that this heightened complexity in prompt exemplar thought steps enhances performance in mathematical problem-solving tasks. This improvement persists regardless of whether a greedy decoding option, which compels the \gls{llm} to select the highest probability answer, or a voting-based bagging approach, such as the self-consistency \gls{cot} method previously presented, is utilized. This method also performs well in mathematical datasets.

Another approach researchers have been taking is to decompose a complex problem into simpler ones. Specifically, the Least-to-Most approach uses an \gls{llm} to decompose a task into sub-tasks \cite{zhou2023leasttomost}. Then, each task is solved sequentially by the \gls{llm}. Every solution generated by this \gls{llm} is given as context to the next question. Least-to-most shows increased performance in symbolic, math and commonsense tasks. 

A different approach known as ``selection-inference'' splits each reasoning step, or thought, into two components: the selection component, which selects a subset of information present at the prompt, and the inference component which produces new content based on the information passed to it by the selection step \cite{creswell2022selectioninference}. By splitting these steps and allowing the \gls{llm} to process them separately, the authors show performance gains over \gls{cot} in commonsense tasks.

Thought propagation is one of the few attempts at using analogical reasoning. Specifically, the authors prompt \glspl{llm} to propose and solve analogous problems related to the input, then reuse the results of analogous problems to generate a solution for the input directly or derive a knowledge-intensive plan for execution to amend the initial solution obtained from scratch \cite{yu2023thought}.

Although most prompt engineering approaches also fit multi-agent environments, specific techniques are proposed for such a scope. Debate~\cite{du2023improving} is a strategy that arguably circumvents some well-known \gls{llm} challenges, such as factual validity and hallucinations. Debate methods work in multiple rounds, with arguments produced in every round being concatenated and shared between participants. Improvements against \gls{cot} were demonstrated in algebraic and logic reasoning tasks. 

K-level reasoning with language models~\cite{zhang2024klevel} is another multi-agent methodology proposed to improve the reasoning capabilities of \gls{llm} in dynamic environments that require decision-making. The method considers \gls{llm}-based agent capable of reasoning about the expected future decisions of other agents. The results showed a trade-off between depth and divergence of such a recursive process, with significant results noticed in multi-agent competitive tasks.
\\ \\
\textit{ii. Fine-Tuning Approaches}
\\ 

OPT-R is an approach that incorporates explanations of answers in fine-tuning of an 
\gls{llm}. The authors found that incorporating explanations in fine-tuning of models yielded significant improvements in math and logical tasks \cite{Alkhamissi_2023}.

Fine-tune-CoT is a method that prompts a very large teacher model, to solve complex questions via zero-shot chain-of-thought reasoning \cite{Ho2022LargeLM}. The authors then use the reasoning samples to fine-tune a much smaller student model. This approach addresses scalability disadvantages of \gls{cot} and its derivatives, which require very large models such as GPT-3 in order to function properly. The authors show that Fine-tune-CoT enables substantial reasoning capability in small models, far outperforming prompt-based baselines and even the teacher model in many tasks.

In addition to \gls{sft}, where \glspl{llm} are trained offline using a labeled dataset and supervised learning algorithms, approaches that use fine-tuned \glspl{llm} for reasoning, also use \gls{rl} algorithms. 

Reasoning with Reinforced Fine-Tuning (REFT) uses a combination of \gls{sft} and \gls{rl}. In a so-called warm-up stage, the \gls{llm} is fine-tuned on a dataset comprising of (user task, \gls{cot}) tuples. In the \gls{rl} stage, the \gls{llm} improves its performance using online self-learning. The model repeatedly samples its own responses, evaluates for correctness and updates its parameters \cite{luong2024reft}. The authors show advances over baseline \gls{cot} approaches in mathematical tasks.
\\ \\
\textit{iii. Neural-Symbolic Approaches}
\\ \\
A third category of approaches involves the combined use of \glspl{llm} and symbolic methods to apply reasoning to solve complex problems. 

One such example is AlphaGeometry, which, given a geometrical problem, uses an \gls{llm} to create symbolic constructs that a symbolic engine can use to reach a solution \cite{Trinh2024}. The approach performs second best on a benchmarking test of 30 Olympiad geometry problems. 

Another example is Logic-LM, which combines \glspl{llm} that translate a natural language problem into a symbolic formulation \cite{pan2023logiclm}. Afterwards, a deterministic symbolic solver performs inference on the formulated problem. The authors also introduce a self-refinement module, which utilizes the symbolic solver's error messages to revise symbolic formalizations. The approach proves more effective than simple \gls{cot} in commonsense tasks.

In another example, CARING combines an \gls{llm}, which is used to represent the knowledge of a problem, and a symbolic solver that performs reasoning using the knowledge of the \gls{llm} \cite{yang2023neurosymbolic}. Specifically, the \gls{llm} transforms knowledge to Prolog language statements such as rules and facts, and then a symbolic solver uses these statements to reason. The approach improves baseline \gls{cot} in math and commonsense, \gls{qa}-type of problems.

Another approach is to consider the \gls{llm} as an agent, capable of interacting with the environment similarly as \gls{rl} agents \cite{fang2024large}. The idea here is that the environment consists of a ``game'' that is the textual description of a task and a symbolic module that helps the agent solve parts of the problem. In this way, the \gls{llm} agent handles the linguistic type of thoughts, while the symbolic thoughts (such as doing mathematical calculations) are delegated to the symbolic module. Results show that the agent outperforms \gls{rl}-based baselines in math, commonsense and symbolic tasks.

\subsubsection{\glspl{gan}}

In the context of reasoning, \glspl{gan} are used to generate missing knowledge that is either essential for reasoning or part of the reasoning itself.

Knowledge Completion \glspl{gan} (KCGANs), attempt to complete missing knowledge in \glspl{kb}, by generating predicates for entity-relation pairs. While discriminative models have already been used in this manner, the advantage of using \glspl{gan} is that they do not require negative samples to learn the distribution of positive samples \cite{ZIA2021543}. As such, they have an advantage over their discriminative counterparts, given that \glspl{kb} does not typically contain negative samples, necessitating the manual generation of those samples, which can prevent the learning of sufficiently robust classifiers.

In the health domain, \glspl{gan} have been used to diagnose cognitive decline from brain scans by hypothesizing how a healthy brain's connections (functional connectivity) might change due to the medical condition \cite{shen2024afbt}. Specifically, a \gls{gan} creates a map highlighting these potential changes using counterfactual explanations and by manipulating the scan data and then uses this map to train a classifier. This approach offers a diagnostic result and reveals which brain regions are most likely affected by cognitive decline. 

\subsubsection{\glspl{vae}}

As with \glspl{gan}, using \glspl{vae} for reasoning is done in conjunction with other methods. 

One such example in the neurosymbolic domain is VAEL, which modifies the generation part of \gls{vae} such that a symbolic part of the latent space is used for probabilistic logic programming \cite{misino2022vael}. This allows the capability of the \gls{vae} to generalize to novel tasks without retraining. The authors demonstrate the applicability of VAEL in commonsense tasks using image generation types of use cases.

Another set of approaches relies on the idea of the conditional \gls{vae} (CVAE), introduced by Sohn et al. \cite{NIPS2015_8d55a249}. The concept behind CVAE revolves around incorporating conditional input to direct the generation process. Specifically, input data is combined with a conditional variable and fed into an encoder, which produces a latent representation capturing pertinent data features based on the condition. Subsequently, the decoder utilizes this latent representation and the same conditional variable to generate an output that aligns with the data and the condition. While the original CVAE paper did not necessarily evaluate performance on reasoning tasks and/or include any reasoning processes behind creating the conditional variable, several newer approaches have described such processes.  

For example, Wang et al. have used a variation of CVAE, called Plan-CVAE, to improve story generation capabilities --- a symbolic reasoning task \cite{wang-etal-2020-plan}. The approach involves using a planner that extracts and expands on a set of keywords given a user input (for example, a story title). The planner combines a keyword extraction algorithm and a \gls{rnn}, specifically a \gls{lstm}, which generates new keywords based on the extracted keywords. Subsequently, the set of keywords is used as a conditional variable on the CVAE. The approach yields better results than \gls{soa} on storytelling aspects such as thematic consistency and wording diversity. 

In another example, Reasoning Skill Discovery (RSD) targets the generation of exemplars for \gls{cot}-type of approaches \cite{xu2023latent}. Typically, exemplar generation is a human task. RSD attempts to extract the reasoning skills required for a specific problem and then select a set of exemplars that demonstrate the required reasoning skills. The approach involves using a CVAE to discover reasoning skills via unsupervised learning. Based on these skills, exemplars are chosen from an example bank of question-rationale pairs to prompt the \gls{llm} using one of the \gls{cot} methods. RSD shows improvements over \gls{soa} in math tasks. 

\subsubsection{Diffusion Models}

In diffusion models, principles of \gls{cot} first described in \glspl{llm} have also been applied in the so-called ``diffusion of thought'' (DoT) \cite{ye2024diffusion}. Although still in an earlier phase than their \gls{llm} counterparts, the approach indicates that diffusion models are simpler, have fewer parameters, and have the potential for better scalability. In math problems, DoT performs comparably to GPT-2. The authors identify the lack of pre-trained foundation diffusion models as a major barrier to achieving higher performance levels.

%% file: sections/section4_telco-uc.tex

\section{Reasoning in Mobile Networks}
\label{section4_telecom_use_cases}

\par{Due to the growing complexity of telecommunication networks, manually monitoring and managing them has become impractical for human operators (see also section \ref{subsec:network_evolution}). Machine reasoning-based approaches that automate decision-making processes, strive to match the cognitive capabilities of human experts while also scaling them for practical implementation.}

    \begin{figure}
    \centering
    \includegraphics[width=1\columnwidth]{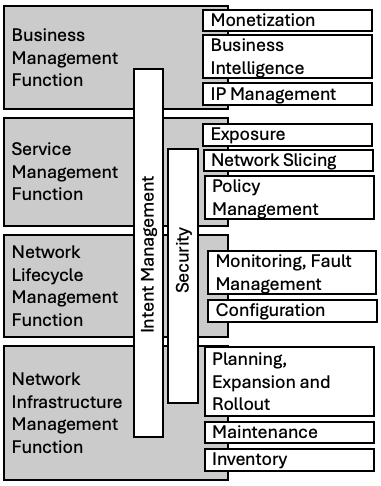}
   \caption{Areas of the mobile network where critical thinking approaches can be of significant assistance.}
    \label{fig:telecom_reasoning}
\end{figure}


Network management and exposure are two areas which can benefit from critical thinking. Insights from works dating back to the 90s demonstrate the applicability of model-based \cite{397221} and rule-based \cite{1508160} approaches to managing telecommunications networks through reasoning. Recent advancements in cognitive networks have advocated for agent-based cognitive frameworks, emphasizing the role of reasoning within the \gls{OSS} of a telecommunication network. In these frameworks, intelligent agents leverage diverse information sources to make operational decisions \cite{8597732, 10243548}. For the rest of this section, we will be focusing on network management-related use cases. 

Figure \ref{fig:telecom_reasoning} illustrates a layered structure of management functions of a mobile telecommunications network. The structure is adapted from \gls{itu} \gls{tmn} layer architecture, with merged ``element'' and ``network element'' layers in a common network infrastructure management layer \cite{ITUTM3010}. 

\begin{itemize}
\item Business management functions include customer-facing use-cases such as monetization and billing, business intelligence and analytics and customer relationship management.
\item Service management functions are use-cases that manage aspects of communication services the mobile network operators provide to their users. These include network slicing and \gls{qos} policy management, security management including Zero-Trust, mobility management, etc.
\item Network life-cycle management functions that include configuration of network equipment, performance monitoring and reporting, inventory, fault and security management.
\item Network infrastructure management functions that include network planning and design, deployment and capacity planning and scaling. 
\end{itemize}

We proceed to review relevant literature and describe each business management layer in greater detail. The reader should note that this is not an exhaustive exploration of all use-cases that \gls{genai} could be applied in, but rather a distillation of those use cases that entail complex problems that could be solved using critical thinking techniques.

\subsection{Business Management}

This category includes use cases that directly impact business operations of a mobile network operator. One group of use cases is monetization and in particular billing process. In previous generations of mobile networks, such as \gls{3g} and \gls{4g}, billing relied on pricing models structured around the projection of user demand for data consumption. However, with \gls{5g} pricing models are set to become more complex, as more information needs to be considered. This information includes service parameters such as \gls{qos}, \gls{qoe}, network parameters such as current resource availability and customer information such as \glspl{sla}. Different types of customers, for example \gls{m2m}, enterprise, and others, create additional complexity \cite{10258042}. Addressing these multifaceted aspects of billing in 5G networks requires sophisticated pricing models, that can adapt dynamically to evolving service requirements and network conditions, thereby presenting a significant challenge for operators and service providers. 

Business intelligence is another area which benefits from use of \gls{ai} \cite{arti1}. Researchers have argued that modern business intelligence needs to incorporate unstructured data from heterogeneous sources into analytics generation, as information systems become more digitized and provide real-time, updated information \cite{10.1007/978-3-642-41924-9_2}. In this environment, reasoning techniques that have the ability to evaluate large unstructured pieces of knowledge and still produce relevant analytics is critical. 

Another area is \gls{ipr} management. Detection of infringement also involves analysis of unstructured data from heterogeneous data sources and correlation of this data to a patent portfolio, also requires advanced techniques and tools such as reasoning. This category of use cases, also includes other intellectual property than patents, such as trademarks. Although still early in telecom, infringement detection systems have already been in use in other domains, such as e-commerce \cite{hu2023tmid}. 

\subsection{Service Management}

This category includes use-cases that manage services rendered by the mobile network. In \gls{5g}, these are predominantly communication services, but in \gls{6g}, service offerings may expand to other types such as positioning and \gls{jcas}. 

Management of network slices\footnote{In a mobile network context, a network slice is defined as a virtualized instance of the network that is customized to meet the specific requirements of an application, service, or user group, enabling efficient resource allocation and optimized performance.} is a popular category for reasoning use-cases. In \gls{soa}, reasoning techniques have been used to connect human management decisions to network slice configuration \cite{9059462}. In another approach reasoning has been applied for dynamic network slice composition based on policies \cite{9059366}. Therefore, critical thinking \gls{ai} techniques in network slicing can serve dual purposes: firstly, for explaining to humans decisions necessary for fulfilling network slice requirements, and secondly, in a reverse manner, for assembling and configuring network slice instances by leveraging various data sources like \glspl{sla} and policies.

Policy management is another group of use cases that include a wide range of functionalities, including \gls{qos} and \gls{qoe} management, traffic prioritization, bandwidth allocation,  access control and traffic steering rules. Making decisions for configuring aspects of those policies is an increasingly complex issue, as it involves use of information from different sources that may include unstructured data and application of logic. In one example, a logic reasoning approach aggregates traffic flows in \gls{qos} classes at the \gls{ran} edge, improving performance over \gls{soa} \cite{9416243}. In another example, a traffic steering solution based on fuzzy logic and reinforcement learning modifies handover parameters to optimize performance against multiple parameters \cite{MUNOZ2014100}.

Exposure concerns these use cases that expose network information to third parties, but also allow third parties to configure aspects of the network. There can be multiple interfaces for such an exposure, for example \gls{3gpp}-standardized \gls{scef} and \gls{nef}\cite{3gpp.29.522}, but also higher-layer interfaces, such as the intent-based network management interfaces defined by \gls{ietf} \cite{jeong-nmrg-ibn-network-management-automation-03}. Reasoning techniques can be applied to translate user domain concepts to network domain concepts and vice versa \cite{9128422}. 

\subsection{Network Lifecycle Management}

Network life-cycle management involves various functions, including monitoring, fault management, and configuration, which are essential for ensuring the seamless operation. Monitoring encompasses real-time observation of network performance and traffic to identify potential issues and ensure optimal functioning. Fault management involves the detection, isolation, and resolution of network faults to minimize downtime and maintain service availability. In one example, case-based reasoning is applied to detect network faults and automatically recover \cite{8598115}.

Configuration management entails the planning, deployment, and maintenance of network device settings and parameters to ensure consistency, security, and compliance with organizational policies and standards. Together, these functions contribute to the effective management and optimization of networks throughout their life-cycle. In one set of examples, semantic reasoning is used in order to semantically model configuration parameters and map those parameters between domains, e.g., the user domain and the network domain \cite{1514535, 7987313}. Typically, as achieving interoperability among configuration management domains entails using knowledge from various domains to translate between network parameters and user domain concepts (e.g., requirements), we believe that employing critical thinking techniques could be beneficial.

\subsection{Network Infrastructure Management}

Mobile network infrastructure management encompasses a range of critical tasks vital for ensuring the seamless operation and growth of mobile networks. Network planning involves strategically designing and optimizing network layouts to meet current and future demands efficiently. Network expansion entails identifying areas where network coverage needs to be extended or improved, often in response to changing user needs or geographic expansion. Network roll-out involves the physical deployment of infrastructure components such as towers, antennas, and base stations to implement planned network expansions. Network maintenance includes field service operations to maintain network equipment. Inventory management involves tracking and managing the various hardware and software components within the network infrastructure. 

For some of these use-cases, reasoning approaches can be used to build explainable \gls{ai} systems that provide explanations in natural language to human operators, accelerating the process and reducing the learning curve for inexperienced personnel \cite{wang2023applications}. In another set of use cases, network can perform complex operations by itself, without the need for human supervision. An example would be the use of reasoning in cognitive radios, for network planning in real time, without the need for human engineers \cite{7368827}. 

\subsection{Overarching Use-Cases}

In addition to use cases belonging to different management functions of the mobile network, as illustrated in figure \ref{fig:telecom_reasoning}, there are some use cases that are overarching, and apply to multiple functions. One such group of use cases is those that use reasoning for intent management, an essential technology towards fully autonomous networks. Intent-driven frameworks typically employ knowledge bases and machine reasoning techniques to decompose higher level intents to lower level representations in a process known as ``intent decomposition'' \cite{9605059}. This process may span multiple layers, starting from the business layer, where user intents get decomposed to network services, and subsequently to network instructions that can be configured directly on operational network equipment. As high level instructions, intents are subject to conflict detection and resolution, a process that can be implemented using adapted algorithms \cite{zheng_intent_2022} or with machine reasoning \cite{baktir_intent-based_2022}.

Another group of use cases is about network security, which also spans multiple management functions. Researchers for example have proposed the creation of a security knowledge graph, based on reasoning techniques, which contains knowledge that can be used to defend against threats across \gls{5g} network layers \cite{9083674}.

\subsection{Use Case Mapping}

In this section, we map the use cases previously presented in section \ref{section4_telecom_use_cases} to various reasoning approaches that were presented in section \ref{section3_relevantApproaches}. It is important to note that this mapping is approximate, as the intricacies of real-world scenarios often defy straightforward categorization. Our aim here is to propose potential avenues for future exploration by researchers rather than to provide a rigid, definitive mapping. 

In scenarios necessitating human interaction, it is conceivable that \glspl{llm} will play a role in soliciting human feedback or documentation specified in some form of natural language. Use cases here may include intent-based autonomous networks, as well as network exposure functions for retrieving information from the network or configuring aspects of the mobile network. 

Another example of interacting with the humans may be assisting in producing and translating documents across domains, such as generating, interpreting and mapping business-domain documents like service agreements into their operational aspects, to be understood by the operations teams according to the operational constraints. This requires making connections across knowledge domains, interpreting, summarizing and translating semi-formal natural language text, and simpler reasoning to ensure that the constraints are met.

In the scope of Intent-based networking~\cite{survey_intent_based_networking2023}, many generative models described so far may find useful applications. Intents are language objects specified in a formal (RDF, UML) or natural language~\cite{3gpp.28.312, tmforum.ig1253}. A natural language intent that reaches the automation infrastructure must sometimes be translated into the expected formal language, a task that suits \glspl{llm} very well. Subsequently, such intents must be broken down and distributed to the proper domains of responsibility. This task requires the capacity to aggregate multiple domain knowledge sources and the reasoning capabilities to decide how to decompose and redistribute the intents. Finally, when low-level intents reach the proper domains, actions must be taken to change the network state and help satisfy the overall constraints and goals specified by the intents~\cite{er_satheesh_marl_2022}. The decision-making aspects related to this last phase of the process can benefit from methodologies enabling planning via trajectory generation (such as conditional diffusion models~\cite{ajay2023is}) and conflict detection (CVAEs~\cite{wang-etal-2020-plan}).

For use cases involving human interaction and content generation, \gls{genai} techniques would benefit from the use of critical reasoning mechanisms. Inductive reasoning techniques would help to generalize outcomes from a few examples. This would be needed for root cause analysis of telecommunication system failures or deriving procedural rules for human participants. Deductive reasoning would enable generating new knowledge artifacts that are not present in the training dataset. Such deductive reasoning techniques would be useful in autonomous network environments, wherein prior knowledge may not be available for all cases. Abductive reasoning would be useful to estimate likely explanations for outcomes, specially when there are no formal guarantees possible. Within intent-driven networks, providing reports and explanations to stakeholders would benefit from abductive reasoning techniques. Finally, analogical reasoning would be useful both for generalizing outcomes (learning from other situations) or for explanation argumentation. Specially for complex network deployments involving multiple hierarchies, geographic locations and business requirements, a combination of the above critical reasoning techniques would be needed. 

In the realm of content generation without human supervision, limitations in producing such content may manifest in the form of symbols, potentially arising from a reasoning process. Such use cases may involve monetization (e.g., generation of billing plans), network planning and expansion (e.g., generation of network topologies and coverage maps) as well as generation of perturbed attack patterns for enhanced security and fraud detection. In this category of use cases, other algorithms than \glspl{llm} may be used, for example CVAEs and \glspl{gan}.

%% file: sections/section5_challenges_opportunities.tex
\section{Challenges and Opportunities} \label{section5_challenges_opportunities}

From our examination of the literature, it becomes apparent that the integration of reasoning with \gls{genai} technologies is a prominent field of study. This observation is underlined by the large volume recently published works, with the vast majority being two years old or less at the time of writing this paper. 

This trend is particularly notable in the context of \glspl{llm}, where strategies like structuring prompts to encourage critical thinking (e.g., \gls{cot} approaches) or retraining existing models with meticulously curated datasets demonstrate comparable, if not superior, performance in certain types of reasoning tasks compared to current state-of-the-art methods. Also, if the problem domain is very specific, they also demonstrate effectiveness as problem solvers, mirroring the thought process they have been prompted or trained with (see \gls{llm} prompt-engineering and fine-tuning sections). 

Nevertheless, it's important to acknowledge that while generative models such as \glspl{llm} can be adapted to emulate reasoning-like behavior, this does not necessarily equate to a true reasoning capability \cite{Kambhampati_2024, lecunn2022}. Ensuring the correctness of purely \gls{genai} based algorithms remains a challenge. However, generative models can be utilized as important components in a complete reasoning process. From one perspective, they excel in converting information from natural language to symbols, to be later processed by a symbolic solver \cite{kambhampati2024llms} (see ``neural-symbolic approaches'' section). Other hybrid approaches include the use of conditions - derived from a reasoning process, to guide content generation (see for example CVAE type of approaches). Recent research also suggests a shift towards a generate-verify paradigm, leveraging content generation abilities of \gls{genai} alongside robust verification tools \cite{valmeekam_planning_2023, silver2024generalized}.

While \glspl{llm} excel in generating language, their ability to produce executable plans, particularly in formalized syntax like Planning Domain Definition Language (PDDL), remains limited \cite{kambhampati2023role}. Integrating \glspl{llm} into planning processes in telecommunications, such as network design and assurance, faces challenges due to these limitations. While \gls{llm}s reason about generating probably ``correct'' facts from text, generating causal relations, formal pre-conditions and effects as well as provenance of reasoning is at a rudimentary stage \cite{hammond2023}. Recent efforts explore leveraging LLMs to enhance plan soundness verification and combining them with Reinforcement Learning (RL) techniques to improve RL agent training in complex telecom use cases \cite{zhou_large_2024, Franceschelli2024}. As presented in \cite{silver2024generalized}, \glspl{llm} may be further used to generalize planning within two aspects: (i) using chain of throught strategies, use \gls{llm}s to create plans and execute them (ii) use corrective re-prompting to generate strategies to avoid plan failures.

%% file: sections/section6_conclusion.tex
\section{Conclusion} \label{section6_conclusion}

In this paper, we have presented an examination of reasoning methodologies concerning \gls{genai} within the context of mobile networks. The impetus for this investigation arises from two main factors: firstly, the notable success of \gls{genai} technologies across diverse problem domains within a brief period, and secondly, the increasing complexity anticipated in future network generations, demanding a heightened level of autonomy that necessitates reasoning-based approaches.

We start by examining the evolution of mobile networks in tandem with advancements in \gls{genai} and highlight the growing role of \gls{genai} and reasoning in assuming critical analytical functions within these networks, concluding with an analysis of the interdependence between reasoning techniques and the quality of structured knowledge.

We continue by presenting a critical thinking task-based taxonomy of generative models, and a set of telecom use cases where reasoning and planning approaches can be applied in. We conclude by consolidating our findings to provide research challenges and opportunities as avenues for future work.